# Near-optimal Nonmyopic Value of Information in Graphical Models

**Andreas Krause**
Carnegie Mellon University

**Carlos Guestrin**
Carnegie Mellon University

## Abstract

A fundamental issue in real-world systems, such as sensor networks, is the selection of observations which most effectively reduce uncertainty. More specifically, we address the long standing problem of nonmyopically selecting the most informative subset of variables in a graphical model. We present the first efficient randomized algorithm providing a constant factor $(1-1/e-\varepsilon)$ approximation guarantee for any $\varepsilon > 0$ with high confidence. The algorithm leverages the theory of *submodular functions*, in combination with a polynomial bound on sample complexity. We furthermore prove that no polynomial time algorithm can provide a constant factor approximation better than $(1-1/e)$ unless $\mathbf{P} = \mathbf{NP}$. Finally, we provide extensive evidence of the effectiveness of our method on two complex real-world datasets.

## 1 Introduction

In decision making, where one can select between several informative but expensive observations, it is often a central issue to decide which variables to observe in order to achieve a most effective increase in expected utility. We consider systems, such as sensor networks, where utility is usually associated with certainty about the measured quantity itself, and the task is to efficiently select the most *informative* subsets of observations. Consider, for example, a temperature monitoring task, where wireless temperature sensors are distributed across a building as shown in Fig. 1(a). Our goal in this example is to become most certain about the temperature distribution, whilst minimizing energy expenditure, a critically constrained resource [3].

Unfortunately, as we show in [15], the problem of selecting the most informative subset of observations is $\mathbf{NP}^{\mathbf{PP}}$-complete, even when the underlying random variables are discrete and their joint probability distribution can be represented as a polytree graphical model (even though inference is efficient in these models). To address this complexity issue, it has been common practice (*c.f.*, [20, 4, 1, 17]) to myopically (greedily) select the most uncertain variable as the next observation, or, equivalently, the set of observations with maximum joint entropy. Unfortunately, these greedy approaches are not associated with formal performance guarantees. Moreover, selecting observations with maximal joint entropy is an *indirect* measure of our value of information goal, the minimization of the remaining uncertainty *after* these observations are made. In sensor networks, for example, we are interested in a *direct* measure of value of information, e.g., minimizing the uncertainty about the unsensed positions. We thus define our observation selection problem as that of selecting a subset $\mathcal{A}$ of the possible variables $\mathcal{V}$ that maximizes *information gain*, $I(\mathcal{V};\mathcal{A}) = H(\mathcal{V}) - H(\mathcal{V} \mid \mathcal{A})$, i.e., decrease in uncertainty about unobserved variables.

Although there is a vast literature on myopic optimization for value of information (*c.f.*, [20, 4, 1]), there has been little prior work on nonmyopic analysis. In [9], a method is proposed to compute the maximum expected utility for specific sets of observations. While their work considers more general objective functions than information gain, they provide only large sample guarantees for the evaluation of a given sequence of observations, and use a heuristic without guarantees to select such sequences. In [15], we present efficient optimal algorithms for selecting subsets and sequences of observations for general objective functions, but these algorithms are restricted to chain graphical models.

In this paper, by leveraging the theory of *submodular functions* [16], we present the first efficient randomized algorithm providing a constant factor $(1-1/e-\varepsilon)$ approximation guarantee for any $\varepsilon > 0$ with high confidence, for any graphical model where inference can be performed efficiently. Our algorithm addresses both the joint entropy criterion, and, under weak assumptions, the information gain criterion. We furthermore prove that no polynomial time algorithm can provide an approximation with a constant factor better than $(1-1/e)$, unless $\mathbf{P} = \mathbf{NP}$. In addition to providing near-optimal informative subset selections, our formal guarantees can be used to automatically derive online performance guarantees for any other algorithm optimizing information gain.

Note that, in [15], we prove that computing the conditional entropy used in our criterion is $\#\mathbf{P}$-complete, even when the underlying distribution is represented by a polytree graphical model. We address this complexity problem in this paper by evaluating our objective function using a sampling algorithm with polynomial sample complexity. The algorithm of Nemhauser *et al.* for maximizing submodular functions [16] can only deal with settings where every element has the same cost. In many practical problems, different observations have different costs. Building on recent constant-factor approximation algorithms for maximizing submodular functions where elements have different costs [18, 14], we extend our approach to problems where possible observations have different costs. Finally, we provide extensive empirical validation of our method on real-world data sets, demonstrating the advantages of our information gain criterion and the effectiveness of our approximation algorithms.

## 2 The value of information problem

In this section, we formalize the problem addressed in this paper: nonmyopic selection of the most informative subset of variables for graphical models. In our example, we want to select the subset of sensors from our deployment indicated in Fig. 1(a) that most effectively decreases expected uncertainty about temperatures in different areas of the lab.

More generally, let $\mathcal{V}$ be the finite set of discrete random variables in our graphical model, and $F : 2^{\mathcal{V}} \rightarrow \mathbb{R}$ be a set function, where $F(\mathcal{A})$ measures the residual uncertainty after we observe $\mathcal{A} \subseteq \mathcal{V}$. In most applications, observations are associated with cost, measuring, for example, the energy required to measure temperatures at particular locations. Given, a cost function $c : 2^{\mathcal{V}} \rightarrow \mathbb{N}$ and a budget $L$, we are interested in computing

$$\mathcal{A}^* = \operatorname*{argmax}_{\mathcal{A} \subseteq \mathcal{V} : c(\mathcal{A}) \leq L} F(\mathcal{A}). \tag{2.1}$$

A basic version of this problem is the *unit cost* case, where every observation has unit cost, $c(\mathcal{A}) = |\mathcal{A}|$, and we are allowed to observe up to $L$ sensors. Apart from the *unit cost* case, we will also present results for the *budgeted* case, where the cost function $c$ is linear, i.e., each variable $X \in \mathcal{V}$ has a positive integer cost $c(X) \in \mathbb{N}$ associated with it, and $c(\mathcal{A}) = \sum_{X \in \mathcal{A}} c(X)$. Note that our approach easily extends to settings where certain variables cannot be observed at all, by setting their cost to a value greater than $L$.

A commonly used criterion for measuring uncertainty is the entropy of a distribution $P : \{x_1, \ldots, x_d\} \rightarrow [0, 1]$,

$$H(P) = -\sum_k P(x_k) \log P(x_k),$$

measuring the number of bits required to encode $\{x_1, \ldots, x_d\}$ [2]. If $\mathcal{A}$ is a set of discrete random variables $\mathcal{A} = \{X_1, \ldots, X_n\}$, then their entropy $H(\mathcal{A})$ is defined as the entropy of their joint distribution. The conditional entropy $H(\mathcal{A} \mid \mathcal{B})$ for two subsets $\mathcal{A}, \mathcal{B} \subseteq \mathcal{V}$ is defined as

$$H(\mathcal{A} \mid \mathcal{B}) = -\sum_{\substack{\mathbf{a} \in \operatorname{dom} \mathcal{A} \\ \mathbf{b} \in \operatorname{dom} \mathcal{B}}} P(\mathbf{a}, \mathbf{b}) \log P(\mathbf{a} \mid \mathbf{b}),$$

measuring the expected uncertainty about variables $\mathcal{A}$ after variables $\mathcal{B}$ are observed.

In practice, a commonly used algorithm for selecting observations is to greedily select the next variable to observe as

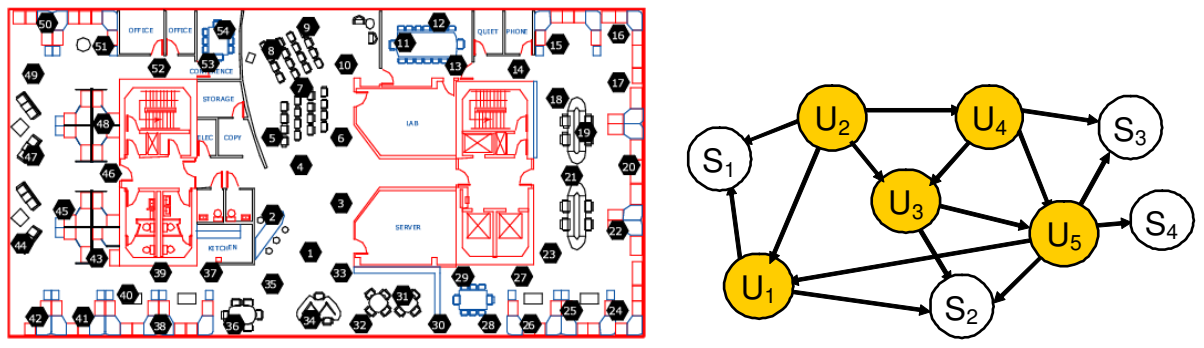

(a) *Sensor network deployment* (b) *Graphical model*

Figure 1: Sensor network deployment.

the most uncertain variable given the ones observed thus far:

$$X_k := \operatorname*{argmax}_X H(X \mid \{X_1, \ldots X_{k-1}\}). \tag{2.2}$$

Using the chain-rule of entropies [2], $H(\mathcal{A} \cup \mathcal{B}) = H(\mathcal{A} \mid \mathcal{B}) + H(\mathcal{B})$, we can decompose the entropy $H(\mathcal{A})$ of a set of variables $\mathcal{A} = \{X_1, \ldots, X_k\}$ as

$$H(\mathcal{A}) = H(X_k \mid X_1, \ldots, X_{k-1}) + \ldots + H(X_2 \mid X_1) + H(X_1),$$

thus, suggesting that the greedy rule in Eq. (2.2) is a heuristic that defines observation *subset selection* task in Eq. (2.1) as the problem of selecting the set of variables that have the maximum joint entropy:

$$\operatorname*{argmax}_{\mathcal{A} : c(\mathcal{A}) \leq L} H(\mathcal{A}). \tag{2.3}$$

It is no surprise that this problem has been tackled with heuristic approaches, since even the unit cost has been shown to be **NP**-hard for multivariate Gaussian distributions [13], and a related formulation has been shown to be $\mathbf{NP}^{\mathbf{PP}}$-hard even for discrete distributions that can be represented by polytree graphical models [15].

A major limitation of this approach is that joint entropy is an indirect measure of information: It aims to maximize the uncertainty about the selected variables, but does not consider prediction quality for unobserved variables. In our sensor networks example, the entropy criterion often leads to the selection of sensors at the border of the sensing field, as sensors that are far apart are most uncertain about each other. Since sensors usually provide most information about the area surrounding them, these border placements "waste" part of their sensing capacity, as noticed in [8].

A more *direct* measure of value of information is the information gain $I(\mathcal{B}; \mathcal{A})$, which is defined as

$$I(\mathcal{B}; \mathcal{A}) = H(\mathcal{B}) - H(\mathcal{B} \mid \mathcal{A}),$$

i.e., the expected reduction in uncertainty over the variables in $\mathcal{B}$ given the observations $\mathcal{A}$. The analogous subset selection problem for the information gain is to compute

$$\operatorname*{argmax}_{\mathcal{A} \subseteq \mathcal{V} : c(\mathcal{A}) \leq L} I(\mathcal{V}; \mathcal{A}). \tag{2.4}$$

For example, we can model our temperature measurement task as indicated in by the graphical model in Fig. 1(b). The temperature in the different rooms is modeled by a set $\mathcal{U} = \{U_1, \ldots, U_n\}$ of hidden nodes, whereas the sensors are modeled as a set of observable nodes $\mathcal{S} = \{S_1, \ldots, S_m\}$. For a set of sensors $\mathcal{A} \subseteq \mathcal{S}$, $H(\mathcal{U} \mid \mathcal{A})$ measures the expected uncertainty about $\mathcal{U}$ given the observations from the sensors in $\mathcal{A}$. Our goal is to define the set of sensor locations that most reduces uncertainty about the hidden variables $\mathcal{U}$.

## 3 Submodularity and value of information

In this section, we derive the *submodularity* property of our objective functions, which will be leveraged by our constant-factor approximation algorithms presented in Sec. 4.

Let $\mathcal{V}$ be a finite set. A set function $F : \mathcal{V} \to \mathbb{R}$ is called *submodular* if it satisfies the "diminishing returns" property,

$$F(\mathcal{A} \cup X) - F(\mathcal{A}) \geq F(\mathcal{A}' \cup X) - F(\mathcal{A}'),$$

for all $\mathcal{A} \subset \mathcal{A}' \subseteq \mathcal{V}$, $X \notin \mathcal{A}$, i.e., adding $X$ to a smaller set helps more than adding it to a larger set. $F(\mathcal{A} \cup X) - F(\mathcal{A})$ is often called the *marginal increase* of $F$ with respect to $X$ and $\mathcal{A}$. The connection between submodularity and value of information is very intuitive in our sensor network example: Acquiring a new observation will help us more when only a few other observations have been made thus far, than when many sensors have already been observed.

The general problem of maximizing submodular functions is $\mathbf{NP}$-hard, by reduction from the max-cover problem, for example. There are branch-and-bound algorithms for maximizing submodular functions, such as the dichotomy algorithm described in [7], but they do not provide guarantees in terms of required running time. Typical problem sizes in practical applications are too large for exact algorithms, necessitating the development of approximation algorithms.

For any submodular function $F$, we can construct a greedy algorithm which iteratively adds element $X$ to set $\mathcal{A}$ such that the marginal increase $F(\mathcal{A} \cup X) - F(\mathcal{A})$ is maximized. Often, we can assume that $F$ is *non-decreasing*, $F(\mathcal{A} \cup X) \geq F(\mathcal{A})$ for all $X$ and $\mathcal{A}$, e.g., observing more variables cannot increase uncertainty. In the context of non-decreasing submodular set functions $F$ where $F(\emptyset) = 0$, and when the unit costs are used for elements of the set, Nemhauser *et al.* [16] prove that the greedy algorithm selects a set of $L$ elements that is at most a constant factor, $(1 - 1/e)$, worse than the optimal set. Using recent extensions of this result to our budgeted setting [18, 14], in this paper, we are able to select observations in graphical models that are at most a constant factor worse than the optimal set.

### 3.1 Submodularity of the joint entropy

The first observation selection criterion we consider is joint entropy $H(\mathcal{A})$ in Eq. (2.3). To prove the submodularity, we must introduce an interesting property of entropy: the "information never hurts" principle [2], $H(X \mid \mathcal{A}) \leq H(X)$, i.e., in expectation, observing $\mathcal{A}$ cannot increase uncertainty about $X$. Since the marginal increase can be written as $F(\mathcal{A} \cup X) - F(\mathcal{A}) = H(X \mid \mathcal{A})$, submodularity is simply a consequence of the information never hurts principle:

$$F(\mathcal{A}\cup X)-F(\mathcal{A})=H(X|\mathcal{A})\geq H(X|\mathcal{A}')=F(\mathcal{A}'\cup X)-F(\mathcal{A}').$$

Submodularity of entropy has been established before [6]. Contrary to the differential entropy, which can be negative, in the discrete case, the entropy $H$ is guaranteed to be non-decreasing, i.e., $F(\mathcal{A} \cup X) - F(\mathcal{A}) = H(X \mid \mathcal{A}) \geq 0$ for all sets $\mathcal{A} \subseteq \mathcal{V}$. Furthermore, $H(\emptyset) = 0$. Hence the approximate maximization result of Nemhauser *et al.*[16] can be used to provide approximation guarantees for observation selection based on the entropy criterion.

### 3.2 Submodularity of the information gain

As discussed in Sec. 2, information gain $F(\mathcal{A}) = I(\mathcal{V}; \mathcal{A})$ is a more direct objective function for observation selection. Again using the information never hurts principle, we have that information gain is a non-decreasing function, and, by definition, that $F(\emptyset) = 0$. Unfortunately, the following counter-example shows that information gain is not submodular in general:

**Example 1.** *Let $X, Y$ be independent boolean random variables with $\Pr[X = 1] = \Pr[Y = 1] = \frac{1}{2}$. Let $Z = X$ **XOR** $Y$. Here, $H(Z) = H(Z \mid X) = H(Z \mid Y) = 1$, but $H(Z \mid X \cup Y) = 0$. Thus, $H(Z \mid X) - H(Z) \geq H(Z \mid X \cup Y) - H(Z \mid Y)$, which implies that information gain, $I(Z; \cdot) = H(Z) - H(Z \mid \cdot)$, is not submodular.* □

Since submodularity is required to use the approximation result of Nemhauser *et al.* [16], their result is not applicable to information gain, in general. Fortunately, under some weak conditional independence assumptions, we prove that information gain is guaranteed to be submodular:

**Proposition 2.** *Let $\mathcal{S}, \mathcal{U}$ be disjoint subsets of $\mathcal{V}$, such that the variables in $\mathcal{S}$ are independent given $\mathcal{U}$. Let* information gain *be $F(\mathcal{A}) = H(\mathcal{U}) - H(\mathcal{U} \setminus \mathcal{A} \mid \mathcal{A})$, where $\mathcal{A} \subseteq \mathcal{W}$, for any $\mathcal{W} \subseteq \mathcal{S} \cup \mathcal{U}$. Then $F$ is submodular and non-decreasing on $\mathcal{W}$, and $F(\emptyset) = 0$.* □

The proofs of all theorems are presented in the Appendix. The assumptions of Proposition 2 are satisfied, e.g., for graphical models with structure similar to Fig. 1(b), where the variables $\mathcal{U}$ form a general graphical model, and the variables in $\mathcal{S}$ each depend on subsets of $\mathcal{U}$. In our sensor network example, we can interpret Proposition 2 in the following way: We want to minimize our uncertainty on the temperatures $\mathcal{U}$, which are measured by noisy sensors $\mathcal{S}$. We can select sensors in $\mathcal{S}$, and potentially make additional, more complicated measurements to estimate certain temperature variables in $\mathcal{U}$ directly (at some, potentially larger, cost).

The following observation identifies the joint entropy as a special case of our setting:

**Corollary 3.** *Let $F(\mathcal{A}) = H(\mathcal{A}) = H(\mathcal{V}) - H(\mathcal{V} \setminus \mathcal{A} \mid \mathcal{A})$. Then $F$ is submodular and non-decreasing on any subset of $\mathcal{V}$, and $F(\emptyset) = 0$.* □

The information gain is another interesting special case:

**Corollary 4.** *Let $\mathcal{S}, \mathcal{U}$ be subsets of $\mathcal{V}$ such that the variables in $\mathcal{S}$ are independent given $\mathcal{U}$. Let $F(\mathcal{A}) = I(\mathcal{U}; \mathcal{A}) = H(\mathcal{U}) - H(\mathcal{U} \mid \mathcal{A})$. Then $F$ is submodular and non-decreasing on $\mathcal{S}$, and $F(\emptyset) = 0$.* □

Note that Corollary 4 applies, for example, to the problem of attribute selection for Naive Bayes classifiers. Hence our algorithms also provide performance guarantees for this important feature selection problem. A convenient property of submodular functions is that they are closed under positive linear combinations. This allows us, for example, to have temperature models for different times of the day, and select sensors that are most informative *on average* for all models.

# 4 Approximation algorithms

In this section, we present approximation algorithms for the unit-cost and budgeted cases, leveraging the submodularity established in Sec. 3.

## 4.1 The unit-cost case

We first consider the unit-cost case, where $c(X_i) = 1$ for all observations $X_i$, and we want to select $L$ observations. This is exactly the setting addressed by Nemhauser *et al.* [16], who prove that the greedy algorithm selects a set that is at most a factor of $1 - (1 - 1/L)^L > (1 - 1/e)$ worse than the optimal set. In our setting, the greedy rule is given by:

**Proposition 5.** *Under the same assumptions as for Proposition 2, when a set $\mathcal{A} \subseteq \mathcal{W}$ has already been chosen, the greedy heuristic for information gain must select the element $X^* \in \mathcal{W}$ which fulfils*

$$X^* \in \underset{X \in \mathcal{W} \setminus \mathcal{A}}{\operatorname{argmax}} \begin{cases} H(X \mid \mathcal{A}) - H(X \mid \mathcal{U}), & \text{for } X \in \mathcal{S}, \\ H(X \mid \mathcal{A}), & \text{otherwise.} \end{cases} \quad \square$$

Proposition 5 has an interesting interpretation: our greedy rule for information gain is very similar to that of the greedy heuristic for maximizing joint entropy, but our information gain criterion is penalized for selecting sensors $X \in \mathcal{S}$ which are "unrelated" to the variables of interest, i.e., those where $H(X \mid \mathcal{U})$ is large.

The greedy algorithm using the rule from Proposition 5 is presented in Algorithm 1, and we summarize our analysis in the following Theorem:

**Theorem 6.** *Algorithm 1 selects a subset $\mathcal{A} \subseteq \mathcal{W}$ of size $L$ such that $F(\mathcal{A}) \geq (1 - 1/e)OPT$, where $OPT$ is the information gain of the optimal set as defined in Eq. (2.4), using $\mathcal{O}(L|\mathcal{W}|)$ computations of conditional entropies.* $\square$

## 4.2 The budgeted case

The result of Nemhauser *et al.* [16] only provides guarantees for the maximization of $F(\mathcal{A})$ for the case where the cost of all observations is the same, and we are simply trying to find the best $L$ observations. Often, different variables have different costs. In our sensor network example, if our sensors are equipped with solar power, sensing during the day time might be cheaper than sensing during the night. In general, the deployment cost for sensors might depend on the

**Input**: $L > 0$, graphical model $G$ for $\mathcal{V} = \mathcal{S} \cup \mathcal{U}$, $\mathcal{W} \subseteq \mathcal{V}$
**Output**: Sensor selection $\mathcal{A} \subseteq \mathcal{W}$
**begin**
    $\mathcal{A} := \emptyset$;
    **for** *j := 1* ***to*** $L$ **do**
        **foreach** $X \in \mathcal{W} \setminus \mathcal{A}$ **do**
            $\Delta_X := H(X \mid \mathcal{A})$;
            **if** $X \in \mathcal{S}$ **then** $\Delta_X := \Delta_X - H(X \mid \mathcal{U})$;
        $X^* := \operatorname{argmax}\{\Delta_X : X \in \mathcal{W} \setminus \mathcal{A}\}$;
        $\mathcal{A} := \mathcal{A} \cup X^*$;
**end**

**Algorithm 1**: Approximation algorithm for the unit cost case.

location. These considerations motivate the consideration of more general cost functions $c(\mathcal{A}) = \sum_{X \in \mathcal{A}} c(X)$, where each variable has a fixed cost $c(X)$.

In [18], a general algorithm has been developed for maximizing non-decreasing submodular functions under a budget constraint with general (additive) costs. Their result builds on an algorithm of Khuller *et al.* [11], who investigated the budgeted MAX-COVER problem (a specific example of a submodular function). Using a partial enumeration technique, the same performance guarantee $(1 - 1/e)$ can be provided, as in the unit-cost case [16]: starting from all possible $d$-element subsets for some constant $d$, a modified greedy heuristic is used to complement these sets. The best such completion is the output of the algorithm. The modified greedy rule for this cost-sensitive case it to select the observation that maximizes the benefit-to-cost ratio: $\frac{F(\mathcal{A} \cup X) - F(\mathcal{A})}{c(X)}$. Since the greedy algorithm starts from every $d$-element subset, the running time of this algorithm is exponential in $d$. Fortunately, as with the specific case addressed by Khuller *et al.* [11], it is sufficient to choose $d = 3$ to provide a $(1 - 1/e)$ approximation guarantee:

**Theorem 7.** *Algorithm 2 selects a subset $\mathcal{A}$ such that $F(\mathcal{A}) \geq (1 - 1/e)OPT$, using $\mathcal{O}(|\mathcal{W}|^{d+2})$ computations of conditional entropies, if $d \geq 3$.* $\square$

The running time of the unit-cost case in Theorem 6 is $\mathcal{O}(|\mathcal{W}|^2)$, while this budgeted case is significantly more expensive ($\mathcal{O}(|\mathcal{W}|^5)$). If we are satisfied with a guarantee of $\frac{(1-1/e)}{2}OPT$, then it is also possible to use a simpler, $\mathcal{O}(|\mathcal{W}|^2)$, algorithm for the budgeted case, which we developed as part of this work [14].

## 4.3 Online guarantees

For additional practical usefulness of our result, note that it can be used to automatically derive an online performance guarantee for any other algorithm optimizing $F$.

**Remark 8.** *Let $F(\mathcal{A})$ be the value computed by our approximation algorithms, and let $F(\mathcal{A}')$ be the value computed by another algorithm. Then we $F(\mathcal{A}')$ is guaranteed to be a $\frac{F(\mathcal{A}')}{F(\mathcal{A})}(1 - 1/e)$ approximation, and we can compute the guarantee at runtime.* $\square$

This observation is useful for the analysis of stochastic ap-

**Input**: $d, L > 0$, graphical model $G$ for $\mathcal{V} = \mathcal{S} \cup \mathcal{U}$, $\mathcal{W} \subseteq \mathcal{V}$
**Output**: Sensor selection $\mathcal{A} \subseteq \mathcal{W}$
**begin**
$\mathcal{A}_1 := \text{argmax}\{F(\mathcal{A}) : \mathcal{A} \subseteq \mathcal{W}, |\mathcal{W}| < d, c(\mathcal{W}) \leq L\}$
$\mathcal{A}_2 := \emptyset$;
**foreach** $\mathcal{G} \subseteq \mathcal{W}, |\mathcal{G}| = d, c(\mathcal{G}) \leq L$ **do**
$\mathcal{W}' := \mathcal{W} \setminus \mathcal{A}$;
**while** $\mathcal{W}' \neq \emptyset$ **do**
**foreach** $X \in \mathcal{W}'$ **do**
$\Delta_X := H(X \mid \mathcal{G})$;
**if** $X \in \mathcal{S}$ **then** $\Delta_X := \Delta_X - H(X \mid \mathcal{U})$;
$X^* := \text{argmax}\{\Delta_X / c(X) : X \in \mathcal{W}'\}$;
**if** $c(\mathcal{G}) + c(X^*) \leq L$ **then** $\mathcal{G} := \mathcal{G} \cup X^*$;
$\mathcal{W}' := \mathcal{W}' \setminus X^*$;
**if** $F(\mathcal{G}) > F(\mathcal{A}_2)$ **then** $\mathcal{A}_2 := \mathcal{G}$
**return** $\underset{\mathcal{A} \in \{\mathcal{A}_1, \mathcal{A}_2\}}{\text{argmax}} F(\mathcal{A})$
**end**

**Algorithm 2**: Approximation algorithm for budgeted case.

proximation algorithms such as simulated annealing [12], which are in certain cases guaranteed to converge to the optimal solution with high probability, but for which one usually does not know whether they have already achieved the optimum or not. Remark 8 provides a lower bound on the quality of approximation achieved and a stopping criterion for these algorithm. Our greedy approximation is also a viable starting point for such local search heuristics.

### 4.4 Hardness of approximation

Our approximation factor $(1 - 1/e)$ may seem loose, but, in this section, we prove that the problem of maximizing the information gain, even in our formulation using conditional independence, cannot be approximated by a constant factor better than $(1 - 1/e)$, unless $\mathbf{P} = \mathbf{NP}$.

We formalize the optimization problem MAXINFOGAIN in the following way: The instances of the problem consist of an integer $L$ and a set $\mathcal{V} = \mathcal{U} \cup \mathcal{S}$ of random variables such that the variables in $\mathcal{S}$ are independent given all $\mathcal{U}$, and a polynomial time Turing machine $M$ for computing $I(\mathcal{U}; \mathcal{A})$ for any subset $\mathcal{A} \subseteq \mathcal{S}$ of size up to $L$. The Turing machine is specified in binary, along with an integer $N$ providing a runtime guarantee. This runtime guarantee is necessary in order to enable the verification of the computations. The solutions comprise the maximum information gain achieved by a subset of size at most $L$.

**Theorem 9.** *The problem MAXINFOGAIN is not approximable within a constant factor better than* $(1 - 1/e)$*, unless* $\mathbf{P} = \mathbf{NP}$. □

Technically, we have to note that computing conditional entropies and hence the actual objective values is in general a $\#\mathbf{P}$-hard problem in its own [15]. Hence, we state our result for problem instances for which these computations can be carried out efficiently, as is the one used in our reduction.

**Input**: $N > 0$, graphical model $G$ for $\mathcal{V}$, $\mathcal{B} \subseteq \mathcal{V}$,
$X \in \mathcal{V} \setminus \mathcal{B}$
**Output**: $\approx H(X \mid \mathcal{B})$
**begin**
$\hat{H} := 0$;
**for** *i = 1* ***to*** $N$ **do**
generate sample $\mathbf{b}$ of $\mathcal{B}$ from $G$;
use prob. inference in $G$ to compute $P(X \mid \mathbf{b})$;
$H(X \mid \mathbf{b}) \leftarrow -\sum_x P(x \mid \mathbf{b}) \log P(x \mid \mathbf{b})$;
$\hat{H} \leftarrow \hat{H} + \frac{1}{N} H(X \mid \mathbf{b})$;
**return** $\hat{H}$;
**end**

**Algorithm 3**: Sampling algorithm for conditional entropy.

## 5 Approximating conditional entropies

The algorithms presented in Sec. 4 require the evaluation of the marginal increases $F(\mathcal{A} \cup X) - F(\mathcal{A})$. Proposition 5 states that this quantity can be computed using one-dimensional conditional entropies. Unfortunately, as we have shown in [15], the computation of conditional entropies is $\#\mathbf{P}$-complete even for discrete polytree graphical models, which motivates the need for approximating these entropies. In this section, we describe a sampling approach only requiring a polynomial number of samples to approximate conditional entropies by an arbitrary $\varepsilon$ difference, with high confidence. Our method in Algorithm 3, which applies to any problem where conditional entropies have to be computed, has sample complexity stated in the following Lemma:

**Lemma 10.** *Algorithm 3 approximates* $H(X \mid \mathcal{B})$ *with absolute error* $\varepsilon$ *and confidence* $1 - \delta$ *using*

$$N = \left\lceil \frac{1}{2} \left( \frac{\log |dom(X)|}{\varepsilon} \right)^2 \log \frac{1}{\delta} \right\rceil$$

*samples, for any* $\varepsilon > 0$ *and* $\delta > 0$. □

In order to use this approximation method for conditional entropies in our observation selection algorithm, we have to ensure that the approximate computation of the marginal increases does not significantly worsen our approximation guarantee. Fortunately, if we can compute the marginal increases $F(\mathcal{A} \cup X) - F(\mathcal{A})$ with an absolute error of at most $\varepsilon$, an argument very similar to the one presented in [16] shows that the greedy algorithm will then provide a solution $\hat{\mathcal{A}}$ such that $F(\hat{\mathcal{A}}) \geq (1 - 1/e)OPT - 2L\varepsilon$ with high confidence. The following Theorem summarizes our analysis of Algorithm 1, using Algorithm 3 for approximate computation of conditional entropies:

**Theorem 11.** *Algorithm 1 computes a set* $\hat{\mathcal{A}}$ *such that*

$$F(\hat{\mathcal{A}}) \geq (1 - 1/e)OPT - \varepsilon,$$

*with probability at least* $1 - \delta$*, in time*

$$\mathcal{O}\left( c_{inf}\ L\ |\mathcal{W}|\ \left\lceil \left( \frac{L \log K}{\varepsilon} \right)^2 \log \frac{L|\mathcal{W}|}{\delta} \right\rceil \right)$$

*for any* $\varepsilon > 0$ *and* $\delta > 0$ *where* $c_{inf}$ *is the cost of probabilistic inference, and* $K = \max_{X \in \mathcal{W}} |dom(X)|$. □

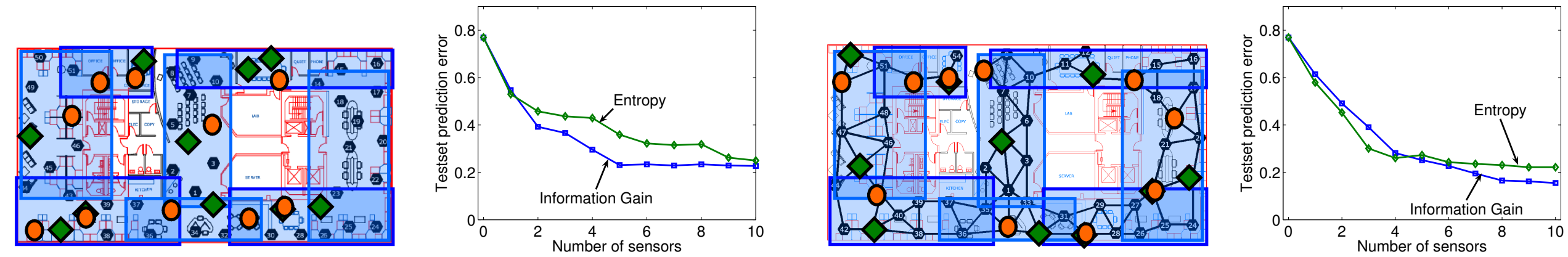

(a) *Placements for hierarc. model* (b) *Prediction error* (c) *Placements for violating model* (d) *Prediction error*

Figure 2: Results for sensor network deployment.

Analogously, the following Theorem summarizes our analysis of Algorithm 2, using the same sampling strategy:

**Theorem 12.** *Algorithm 2 computes a set* $\hat{\mathcal{A}}$ *such that*

$$F(\hat{\mathcal{A}}) \geq (1 - 1/e)OPT - \varepsilon$$

*with probability at least* $1 - \delta$ *in time*

$$\mathcal{O}\left(c_{inf}\; |\mathcal{W}|^{d+2} \left[\left(\frac{L \log K}{\varepsilon}\right)^2 \log \frac{|\mathcal{W}|^{d+2}}{\delta}\right]\right)$$

*for any* $\varepsilon > 0$ *and* $\delta > 0$ *where* $c_{inf}$ *is the cost of probabilistic inference and* $K = \max_{X \in \mathcal{W}} |dom(X)|$, *if* $d \geq 3$. □

Note that, in order to guarantee polynomial running time for Algorithms 1 and 2, the cost $c_{inf}$ of sampling from the graphical model and the probabilistic inference step in Algorithm 3 has to be polynomial. For the case of discrete polytrees, for which the exact maximization is $\mathbf{NP^{PP}}$-hard, however, efficient sampling and inference are possible, as is for all graphs of bounded treewidth.

# 6 Experimental results

## 6.1 Temperature data

In our first set of experiments, we analyzed temperature measurements from the sensor network as mentioned in our running example. We fit a hierarchical model to the data, as indicated in Fig. 2(a). The overlapping rectangles correspond to larger areas of the research lab, where each region is associated with a random variable representing the mean of all sensors contained in this region. Since our regions overlap, sensors may be in more than one region. The sensors are assumed to be conditionally independent given the values of all aggregate nodes.

Our training data consisted of samples collected at 30 sec. intervals from 6 am till 8 pm on five consecutive days (starting Feb. 28th 2004). The test data consisted of the corresponding samples on the two following days. Data was discretized into five bins of three degrees Celsius each. We used our approximation algorithms to learn informative subsets of the original sensors. Fig. 2(a) shows the sensors selected by the information gain criterion (circles) and by the entropy criterion (diamonds) for the unit-cost optimization with $L = 10$. Fig. 2(b) compares the prediction quality (measured as misclassification errors) on the aggregate nodes for both subset selections, using varying subset sizes. The prediction quality of the subset selected by our information gain criterion is significantly better than that of the entropy criterion. Using only five sensors, information gain achieved a prediction accuracy which was not met by the entropy criterion even for subset sizes up to 20 sensors.

To verify whether the algorithm for maximizing the information gain achieves viable results even in the case where its assumptions of conditional independence are violated, we performed another experiment. In addition to the identical hierarchical structure, our new graphical model connects adjacent sensors as indicated in Fig. 2(c), which also shows ten sensor subsets selected using the entropy and information gain criteria. Fig. 2(d) presents the test set prediction errors for both criteria. It can be seen that, in comparison to Fig. 2(b), the prediction accuracy is not much improved by this more complex model, though our information gain criterion manages to decrease the prediction error to below 20 percent, which posed a hard barrier for the basic model.

Fig. 3(a) presents the time required for computing the greedy update steps for a varying number of samples, both for the basic and complex model. As predicted in Theorem 6, the run time is linear in the number of samples, the total number of sensors and the size of the selected subset. To assess how the number of samples affects the performance of Algorithm 1, Fig. 3(b) displays the information gain computed for five element subsets for a varying number of samples. There is no significant increase of the mean information gain, suggesting that the bounds presented in Theorem 10 are very loose for practical applications.

## 6.2 Highway traffic data

In our second set of experiments, we analyzed highway traffic from the San Francisco Bay area [19]. Each detector station, 77 in total, computed aggregated measurements over 5 minutes, reporting the total number of vehicle miles traveled divided by the total vehicle hours for its region.

As for the the temperature data, we fit a hierarchical model, shown in Fig. 3(c). The bay area was logically segmented into overlapping regions, covering bridges and major highway segments. The diamonds represent the aggregate nodes, and the detector stations (squares) are assumed to be conditionally independent given all aggregate nodes. The data was collected over 18 days, of which the last two days were chosen for testing prediction accuracy. We chose a discretization into five bins: below 50 mph, between 50-60, 60-65, 65-70 and more than 70 mph, and the task was to predict the min-

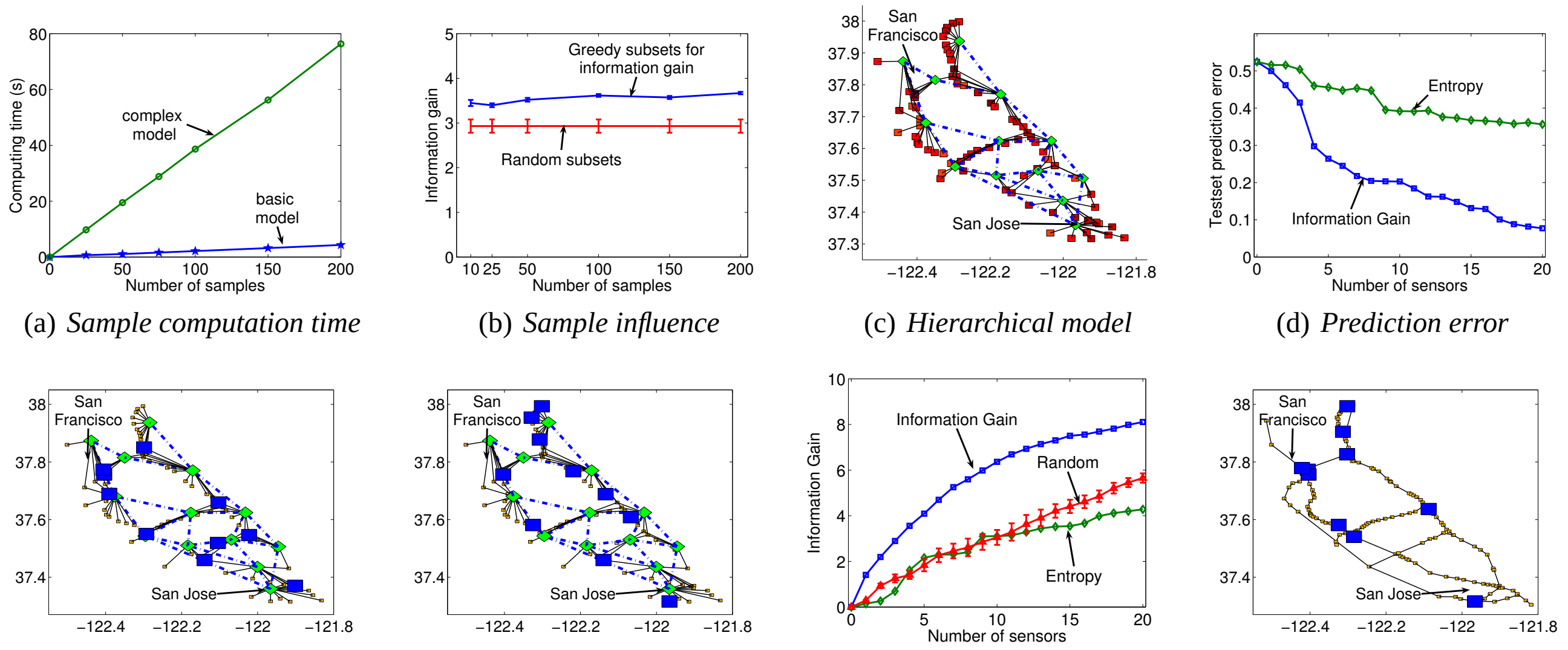

(a) *Sample computation time* (b) *Sample influence* (c) *Hierarchical model* (d) *Prediction error*

(e) *Sensor selection (inf. gain)* (f) *Sensor selection (entropy)* (g) *Objective values* (h) *Second highway model*

Figure 3: Results for sensor network deployment and highway traffic data.

imum speed over all detectors within each region. Fig. 3(d) compares the prediction accuracy for subsets selected using the entropy and information gain criteria. Information gain significantly outperforms the maximum entropy selection. Figures 3(e) and 3(f) show the selected sensor locations for information gain and entropy.

In another experiment, we compared the information gain objective values computed for subsets selected using the greedy heuristic for both information gain and entropy criteria, and random selections; Fig. 3(g) presents these results. Interestingly, the average information gain for randomly chosen subsets is even slightly higher than using the greedy entropy maximization. This indicates that maximizing entropy and information gain appear to be conflicting goals, resulting in qualitatively different behaviors.

Analogously to the temperature data, we performed another experiment with a different model. Instead of a hierarchical model satisfying the conditional independence assumptions for maximization of the information gain, we fit a larger Bayesian network for which the 153 detector stations along the highways were connected, as indicated in Fig. 3(h). This figure also indicates the placement of the most informative sensors, which are concentrated on very similar locations as in the hierarchical model (*c.f.*, Fig. 3(f)).

## 7 Conclusions

We presented efficient randomized approximation algorithms for the long standing problem of nonmyopically selecting most informative subsets of variables in graphical models. Our algorithms provide a constant factor $(1-1/e-\varepsilon)$ performance guarantee with high confidence, both in the unit-cost and in the budgeted case. Our methods can also be used to derive online performance guarantees for other heuristics. The analysis of the algorithms leveraged the concept of submodular functions, and polynomial bounds on sample complexity. We also showed that $(1-1/e)$ is the best performance guarantee possible unless $\mathbf{P} = \mathbf{NP}$. Finally, our empirical results demonstrate the practical applicability of our method to real-world problems: relating the maximization of our submodular objective functions to improving prediction accuracy, and showing the superiority of our information gain criterion to entropy maximization. We believe that our strong guarantees for optimizing value of information in graphical models will significantly improve the handling of uncertainty in large-scale resource-constrained systems, such as sensor networks.

## Acknowledgements

We would like to thank Tommi Jaakkola, Brigham Anderson and Andrew Moore for helpful comments and discussions. This work was partially supported by a gift from Intel Corporation.

## Appendix

*Proof of Proposition 2.* Let $\mathcal{C} = \mathcal{A} \cup \mathcal{B}$ such that $\mathcal{A} \subseteq \mathcal{S}$ and $\mathcal{B} \subseteq \mathcal{U}$. Then

$$H(\mathcal{U} \setminus \mathcal{C} \mid \mathcal{C}) = H(\mathcal{U} \setminus \mathcal{B} \mid \mathcal{A} \cup \mathcal{B}) = H(\mathcal{U} \cup \mathcal{A}) - H(\mathcal{A} \cup \mathcal{B})$$
$$= H(\mathcal{A} \mid \mathcal{U}) + H(\mathcal{U}) - H(\mathcal{C}) = H(\mathcal{U}) - H(\mathcal{C}) + \sum_{Y \in \mathcal{C} \cap \mathcal{S}} H(Y \mid \mathcal{U})$$

using the chain rule for the joint entropy. $H(\mathcal{U})$ is constant, $H(\mathcal{C})$ is submodular and $\sum_{Y \in \mathcal{C} \cap \mathcal{S}} H(Y \mid \mathcal{U})$ is linear in $\mathcal{C}$ on $\mathcal{U} \cup \mathcal{S}$ and hence on $\mathcal{W}$.

To show that $F$ is non-decreasing we consider $\Delta = F(\mathcal{C} \cup X) - F(\mathcal{C}) = H(\mathcal{C} \cup X) - H(\mathcal{C}) + \sum_{Y \in (\mathcal{C} \cup X) \cap \mathcal{S}} H(Y \mid \mathcal{U}) - \sum_{Y \in \mathcal{C} \cap \mathcal{S}} H(Y \mid \mathcal{U})$. If $X \in \mathcal{U}$, then $\Delta = H(\mathcal{C} \cup X) - H(\mathcal{C}) = H(X \mid \mathcal{C}) \geq 0$. If $X \in \mathcal{S}$, then $\Delta = H(X \mid \mathcal{C}) - H(X \mid \mathcal{U}) \geq H(X \mid \mathcal{C} \cup \mathcal{U}) - H(X \mid \mathcal{U}) = 0$ using conditional independence and the fact that conditioning never increases the entropy. $\square$

*Proof of Proposition 5.* We compute the marginal increase $\Delta = F(\mathcal{C} \cup X) - F(\mathcal{C})$ as in the proof of Proposition 2. If $X \in \mathcal{U}$, then $\Delta = H(X \mid \mathcal{C})$ and if $X \in \mathcal{S}$, then $\Delta = H(X \mid \mathcal{C}) - H(X \mid \mathcal{U})$. □

*Proof of Theorem 9.* We will reduce MAX-COVER to maximizing the information gain. MAX-COVER is the problem of finding a collection of $L$ sets such their union contains the maximum number of elements. In [5], it is shown that MAX-COVER cannot be approximated by a factor better than $(1 - 1/e)$ unless $\mathbf{P} = \mathbf{NP}$. Our reduction generates a Turing machine, with a polynomial runtime guarantee, which computes information gains for variable selections.

Let $\mathcal{U} = \{X_1, \ldots, X_n\}$ be a set, and $U_1, \ldots, U_m$ be a collection of subsets of $\mathcal{U}$, defining an instance of MAX-COVER . Interpret $\mathcal{U}$ as a set of independent binary random variables $X_j$ with $\Pr(X_j = 1) = 0.5$. For each $1 \le i \le m$ let $Y_i$ be a random vector $Y_i = (X_{i_1}, \ldots, X_{i_l})$ where $U_i = \{X_{i_1}, \ldots, X_{i_l}\}$, and let $\mathcal{S} = \{Y_1, \ldots, Y_m\}$. Considered stochastically, the $Y_1, \ldots, Y_m$ are conditionally independent given all $X_i$. Furthermore, for $\mathcal{B} \subseteq \mathcal{U}, \mathcal{A} \subseteq \mathcal{S}$ it holds that $H(\mathcal{B} \mid \mathcal{A}) = \sum_{X \in \mathcal{B}} H(X \mid \mathcal{A})$, and $H(X \mid \mathcal{A}) = 0$ iff $X \in \mathcal{A}$, 1 otherwise. Hence if $I \subseteq \{1, \ldots, m\}$, then $H(\mathcal{U}) - H(\mathcal{U} \mid \bigcup_{i \in I} Y_i) = |\bigcup_{i \in I} U_i|$. Thus if we can approximate the maximum of $I(\mathcal{U}; \mathcal{A})$ over all $L$-element subsets $\mathcal{A} \subseteq \mathcal{S}$ by a constant factor of $(1 - 1/e + \varepsilon)$ for some $\varepsilon > 0$, then we can approximate the solution to MAX-COVER by the same factor. Note that the entropies computed in our reduction are all integer-valued, and it is possible to efficiently construct a polynomial time Turing machine computing $I(\mathcal{U}; \mathcal{A})$, with a runtime guarantee. □

*Proof of Lemma 10.* We will approximate the marginal increase $\Delta = F(\mathcal{A} \cup X) - F(\mathcal{A})$ by sampling. Hence we need to compute the number $N$ of samples we need to guarantee that the sample mean $\Delta_N$ does not differ from $\Delta$ by more than $\varepsilon/L$ with a confidence of at least $1 - \frac{\delta}{Ln}$. First we note that $0 \le \Delta \le H(X) \le \log |dom(X)|$. Hoeffding's inequality [10] states that $\Pr[|\Delta_N - \Delta| \ge \varepsilon/L] \le 2\exp(-2N(\frac{\varepsilon}{L \log |dom(X)|})^2)$. This quantity is bounded by $\delta/(Ln)$ if $N \ge \frac{1}{2}(\frac{L \log |dom(X)|}{\varepsilon})^2 \log \frac{2Ln}{\delta}$. □

*Proof of Theorem 11.* Let $\varepsilon, \delta > 0$. We will approximate $F$ by sampling. Let $n = |S|$. In each step of the greedy algorithm, $2n$ sample approximations have to be made. If we run $L$ steps, we have to guarantee a confidence of $1 - \frac{\delta}{2Ln}$ for each sample approximation to guarantee an overall confidence of $1 - \delta$ by the union bound. The result follows from Lemma 10 and the remark on approximate maximization of submodular functions. □

*Proof of Theorem 12.* Let $\varepsilon, \delta > 0$ and $n = |S|$. For the initial partial enumeration, $2dn^d$ conditional entropies have to be computed. For the computation of $\mathcal{A}_2$, the greedy algorithm is run $\mathcal{O}(n^d)$ times. In each step of the greedy algorithm, $2n$ sample approximations have to be made. Since the budget might not be exhausted even for the total set $\mathcal{W}$, there are at most $n - d$ steps. Hence, in total, $\mathcal{O}(n^{d+2})$ sample approximations have to be made, each with an absolute error bounded by $\varepsilon/L$, and a confidence of $1 - \frac{\delta}{4n^{d+2}}$, in order to guarantee an overall error of $\varepsilon$ and confidence of $1 - \delta$ by the union bound. The result follows from Lemma 10 and our result about approximate budgeted maximization of submodular functions [14]. □